\documentclass[11pt]{article}
\usepackage{acl}
\usepackage{times}
\usepackage{latexsym}
\usepackage{amsmath} % assumes amsmath package installed
\usepackage{amssymb}  % assumes amsmath package installed
\usepackage{mathtools}
\usepackage{graphics}
\usepackage{subcaption}
\usepackage{amsmath,amsfonts,bm}

\def\eqref#1{equation~\ref{#1}}
\def\1{\bm{1}}

\DeclareMathAlphabet{\mathsfit}{\encodingdefault}{\sfdefault}{m}{sl}
\SetMathAlphabet{\mathsfit}{bold}{\encodingdefault}{\sfdefault}{bx}{n}

\usepackage{multirow}
\usepackage{hyperref}
\usepackage{xcolor}
\usepackage{url}
\usepackage{setspace}
\usepackage{algpseudocode}
\usepackage{algorithm}
\usepackage{booktabs}
\usepackage{enumitem}
\usepackage[most]{tcolorbox}
\usepackage{textcomp}
\usepackage[table]{xcolor}

\definecolor{TableGreen}{RGB}{0, 196, 0 }

\title{Dual-Uncertainty Guided Policy Learning for Multimodal Reasoning}

\author{
Rui Liu$^{1,2}$,
Dian Yu$^{1}$,
Tong Zheng$^{2}$,
Runpeng Dai$^{3}$,
Zongxia Li$^{2}$, 
Wenhao Yu$^{1}$, \\
\textbf{Zhenwen Liang}$^{1}$,
\textbf{Linfeng Song}$^{1}$,
\textbf{Haitao Mi}$^{1}$,
\textbf{Pratap Tokekar}$^{2}$,
\textbf{Dong Yu}$^{1}$ \\
$^{1}$Tencent Hunyuan, Bellevue \\
$^{2}$University of Maryland, College Park \\
$^{3}$University of North Carolina, Chapel Hill \\
}

\begin{document}

\maketitle

% \vspace{-50pt}

\begin{abstract}
\vspace{-5pt}
% Reinforcement learning with verifiable rewards (RLVR) has advanced reasoning capabilities in multimodal large language models. However, existing methods typically treat visual inputs as deterministic, overlooking the perceptual ambiguity inherent to the visual modality. Consequently, they fail to distinguish whether a model's uncertainty stems from complex reasoning or ambiguous perception, preventing the targeted allocation of exploration or learning signals. To address this gap, we introduce DUPL, a dual-uncertainty guided policy learning approach for multimodal RLVR that quantifies and leverages both perceptual uncertainty (via symmetric KL divergence) and output uncertainty (via policy entropy) to guide policy updates. By establishing an uncertainty-driven feedback loop and employing a dynamic branch prioritization mechanism, DUPL recalibrates the policy advantage to focus learning on states with high perceptual or decisional ambiguity, enabling effective targeted exploration beyond passive data augmentation. Implemented on top of GRPO and evaluated on six multimodal mathematical and general-domain reasoning benchmarks, DUPL improves Qwen2.5-VL 3B and 7B models, achieving accuracy gains of up to 11.2\% on visual math tasks and up to 7.1\% on general-domain reasoning tasks, while consistently outperforming GRPO. These results demonstrate that dual-uncertainty guided policy learning is an effective and generalizable approach for multimodal RLVR.

Reinforcement learning with verifiable rewards (RLVR) has advanced reasoning capabilities in multimodal large language models. However, existing methods typically treat visual inputs as deterministic, overlooking the perceptual ambiguity inherent to the visual modality. Consequently, they fail to distinguish whether a model's uncertainty stems from complex reasoning or ambiguous perception, preventing the targeted allocation of exploration or learning signals. To address this gap, we introduce \textbf{DUPL}, a dual-uncertainty guided policy learning approach for multimodal RLVR that quantifies and leverages both perceptual uncertainty (via symmetric KL divergence) and output uncertainty (via policy entropy) to guide policy updates. By establishing an uncertainty-driven feedback loop and employing a dynamic branch prioritization mechanism, DUPL recalibrates the policy advantage to focus learning on states with high perceptual or decisional ambiguity, enabling effective targeted exploration beyond passive data augmentation. Evaluated on diverse multimodal reasoning benchmarks spanning mathematical and general domains, DUPL achieves solid gains. It improves Qwen2.5-VL accuracy by up to \textbf{12.3\%} (3B) and \textbf{7.9\%} (7B), and Qwen3-VL-Instruct by up to \textbf{10.7\%} (4B) and \textbf{12.4\%} (8B), consistently outperforming GRPO, while seamlessly generalizing to alternative algorithms (DAPO, \textbf{+6.5\%} avg) and architectures (LLaVA-OneVision-1.5, \textbf{+4.7\%} avg). These results demonstrate that DUPL is an effective and generalizable approach for multimodal RLVR.

% Furthermore, we demonstrate DUPL's strong generalizability across architectures and RL algorithms: it effectively enhances the LLaVA-OneVision-1.5-8B model by an average of 4.7% and successfully integrates with DAPO, yielding a 6.5% average improvement over the base model. These results confirm that dual-uncertainty guided policy learning is a highly effective and robust approach for multimodal RLVR.

\end{abstract}

\section{Introduction}
\vspace{-5pt}
\begin{figure*}[ht]
    \centering

    \includegraphics[width=0.82\textwidth]{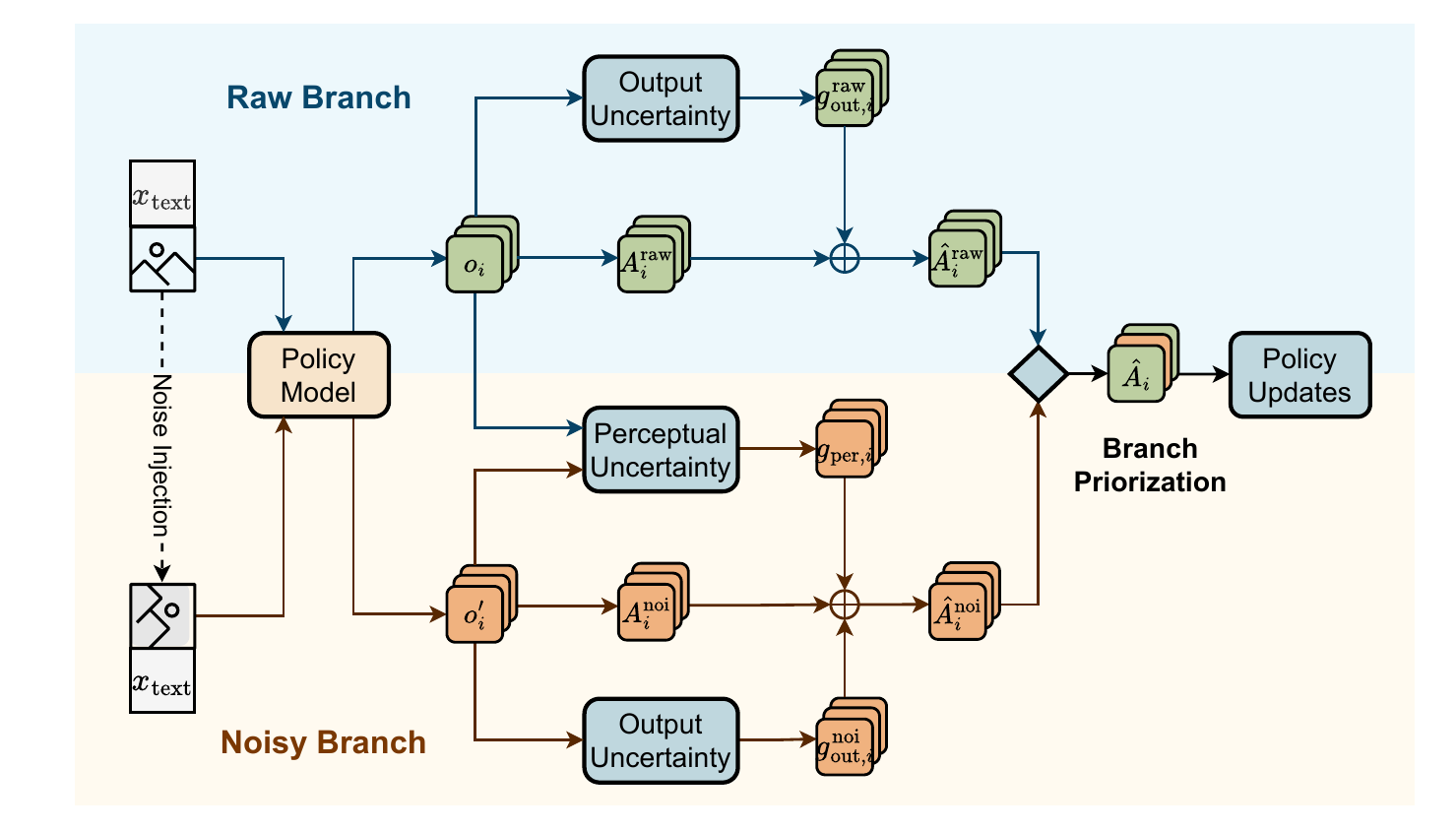}
    \vspace{-8pt}

    \caption{\textbf{Overview of DUPL for multimodal reasoning.} DUPL establishes a dual-uncertainty guided feedback loop for policy updates, capturing both perceptual and output uncertainty. The raw branch processes the original input, while the noisy branch processes a perturbed view. Perceptual uncertainty is quantified via token-level symmetric KL divergence between branches, producing the guidance signal $g_{\text{per}}$ to shape the advantage. Output uncertainty is measured by policy entropy, generating guidance signals $g_{\text{out}}^{\text{raw}}$ and $g_{\text{out}}^{\text{noi}}$ for the raw and noisy branches, respectively. A dynamic branch prioritization mechanism emphasizes uncertainty-driven exploration early in training and gradually shifts focus to the raw branch as learning stabilizes.}

    \label{fig:app}
    \vspace{-12pt}
\end{figure*}

Reinforcement learning with verifiable rewards (RLVR) has substantially improved the reasoning abilities of large language models (LLMs) by optimizing against ground-truth answers~\citep{luong2024reft,lambert2024t,guo2025deepseek,su2025crossing, zheng2025parallel}. However, this outcome-centric approach often suppresses trajectories with valid intermediate reasoning steps that conclude with incorrect final answers, restricting exploration and yielding brittle policies~\citep{dai2025cde}. While text-only strategies mitigate this via uncertainty-aware objectives~\citep{cheng2025reasoning}, diversity-promoting rewards~\citep{li2025jointly}, Pass@k rewards~\citep{chen2025pass,walder2025pass}, intermediate feedback~\citep{setlur2024rewarding}, and entropy regularization \citep{cheng2025reasoning, zhang2025entropy, cui2025entropy, wang2025beyond}, these mechanisms operate strictly within the text or action space. As the paradigm shifts to Multimodal Large Language Models (MLLMs)~\citep{huang2025vision, tan2025reason, peng2025lmm}, where textual reasoning is coupled with complex visual inputs, effective exploration becomes fundamentally more challenging and remains largely underexplored.

% This limitation has been recognized in text-only domains, with mitigation strategies including uncertainty-aware objectives~\citep{cheng2025reasoning}, diversity-promoting rewards~\citep{li2025jointly}, pass@k rewards~\citep{chen2025pass,walder2025pass}, intermediate feedback~\citep{setlur2024rewarding}, and entropy regularization \citep{cheng2025reasoning, zhang2025entropy, cui2025entropy, wang2025beyond}. Notably, most existing methods regulate policy exploration through mechanisms defined purely in the text or action space. As the paradigm expands to Multimodal Large Language Models (MLLMs) \citep{huang2025vision, tan2025reason, peng2025lmm, liu2025stable}, where textual reasoning is coupled with complex visual inputs, effective exploration becomes fundamentally more challenging and remains largely underexplored. 

Current multimodal RLVR approaches typically treat visual input as a fixed, deterministic condition. This overlooks the inherent ambiguity of the visual modality, where ambiguous objects or multiple valid interpretations may exist. While recent approaches~\citep{liu2025noisyrollout,yao2025r1} attempt to improve robustness via visual noise injection during rollout collection, these methods primarily rely on passive data augmentation that leaves the learning objective unchanged, resulting in undirected exploration rather than targeted exploration toward states of genuine uncertainty. Similarly, while perception-aware optimization improves visual grounding~\citep{wang2025perception}, it lacks a mechanism to model or leverage uncertainty to guide policy exploration.

% However, this perceptual determinism is flawed, as it overlooks a key source of ambiguity: the visual modality itself. An image may contain ambiguous objects, or be subject to multiple valid interpretations. While recent approaches~\citep{liu2025noisyrollout,yao2025r1} attempt to encourage exploration and improve visual robustness by injecting noise into images during rollout collection to increase trajectory diversity. However, these methods primarily rely on passive data augmentation and leave the learning objective unchanged, resulting in global and undirected exploration rather than targeted exploration toward states with genuine uncertainty. Similarly, \citet{wang2025perception} make policy optimization perception-aware to improve visual grounding, but do not model uncertainty or use it to guide exploration.

% Taken together, prior work either promotes exploration in the action or text space, or improves data diversity through passive augmentation. This reveals a fundamental limitation in current multimodal RLVR methods: exploration strategies are blind to the source of uncertainty. Existing methods cannot distinguish whether a model's uncertainty stems from complex reasoning or ambiguous perception, and thus cannot allocate exploration or learning signals accordingly. What is missing is a policy learning mechanism that can identify a model's uncertainty and respond in a targeted manner.

This reveals a fundamental limitation in current multimodal RLVR: exploration strategies are \textbf{blind to the source of uncertainty}. By failing to distinguish whether a model’s uncertainty stems from complex reasoning or ambiguous perception, existing methods cannot prioritize learning signals where they are most needed. Consequently, exploration remains inefficiently distributed. What is missing is a policy learning mechanism capable of \textbf{decoupling these uncertainty streams} to enable targeted, uncertainty-aware updates.

To address this gap, we introduce \textbf{DUPL}, a targeted exploration approach for multimodal RLVR through \textbf{D}ual-\textbf{U}ncertainty guided \textbf{P}olicy \textbf{L}earning, as illustrated in Figure \ref{fig:app}. Inspired by closed-loop control principles \citep{hjalmarsson2005experiment}, DUPL forms an uncertainty-driven feedback loop where model uncertainty serves as a measured feedback signal to actively regulate policy updates. Rather than encouraging exploration through indiscriminate, passive data augmentation \citep{liu2025noisyrollout, yao2025r1}, DUPL explicitly explores where the model is uncertain, capturing both perceptual and output uncertainty. Specifically, we transform visual perturbations into an \textbf{active sensitivity probe}: for each training example, we perform a dual-branch forward pass on the original image and a perturbed view, quantifying \textbf{perceptual uncertainty} via the symmetric KL divergence between the induced policy distributions. In parallel, DUPL models \textbf{output uncertainty} via the policy’s entropy in the action space. By jointly leveraging these two uncertainties as guidance signals to shape advantage for policy updates, DUPL dynamically steers exploration toward perceptually and decision-wise ambiguous states. Furthermore, we employ a dynamic branch prioritization mechanism that emphasizes uncertainty-driven exploration early in training before shifting focus to the original view as learning stabilizes. 

To validate DUPL, we evaluate its performance across diverse mathematical and general-domain reasoning benchmarks: MathVerse~\citep{zhang2024mathverse}, MathVista~\citep{lu2023mathvista}, WeMath~\citep{qiao2024we}, HallusionBench~\citep{guan2024hallusionbench}, ChartQA~\citep{masry2022chartqa}, LogicVista~\citep{xiao2024logicvista}, ChartMuseum \citep{tang2026chartmuseum}, MMReason \citep{yao2025mmreason}, and VisuLogic \citep{xu2025visulogic}. When trained on the MMRL30k dataset~\citep{zhu2025shuffle}, DUPL improves Qwen2.5-VL \citep{bai2025qwen25vltechnicalreport} accuracy by up to 12.3\% (Avg.\ 8.7\%) at the 3B scale and 7.9\% (Avg.\ 5.8\%) at the 7B scale (Table \ref{tab:main}), while enhancing Qwen3-VL-Instruct by up to 10.7\% (Avg. 5.3\%) for the 4B model and up to 12.4\% (Avg. 6.3\%) for the 8B model (Table \ref{tab:qwen3}), consistently outperforming GRPO \citep{shao2024deepseekmath}. DUPL also demonstrates strong generalization to other RL algorithms: when trained with DAPO~\citep{yu2026dapo}, it improves the 7B base model by an average of 6.5\% and surpasses the DAPO baseline (Table~\ref{tab:dupl_dapo}). Moreover, applying DUPL to an alternative base model, LLaVA-OneVision-1.5-8B-Instruct~\citep{an2025llava}, yields an average improvement of 4.7\% (Table~\ref{tab:llava}). In summary, the core contributions of our work are as follows: 

\begin{itemize}[left=0pt]
\vspace{-8pt}
\item We identify a core limitation in multimodal RLVR: the inability to separate perceptual ambiguity from reasoning uncertainty, causing inefficient, undirected exploration.

\vspace{-5pt}
\item We introduce DUPL, a policy learning framework that decouples and quantifies perceptual and output uncertainty to recalibrate advantage through an uncertainty-driven feedback loop.

\vspace{-5pt}
\item We propose an active probing mechanism with dynamic branch prioritization, transforming visual perturbations from passive augmentation into a principled signal for targeted, uncertainty-aware optimization.

\vspace{-5pt}
\item Through extensive evaluation, DUPL achieves solid gains. It improves Qwen2.5-VL accuracy by up to \textbf{12.3\%} (3B) and \textbf{7.9\%} (7B), and Qwen3-VL-Instruct by up to \textbf{10.7\%} (4B) and \textbf{12.4\%} (8B), outperforming GRPO, while seamlessly generalizing to alternative algorithms (DAPO, \textbf{+6.5\%} avg) and architectures (LLaVA-OneVision-1.5, \textbf{+4.7\%} avg).
\end{itemize}

\vspace{-10pt}
\section{Approach} \label{sec: app}
\vspace{-5pt}
% We build upon Group Relative Policy Optimization (GRPO) \citep{shao2024deepseekmath} as the underlying RL algorithm in this work (see Appendix \ref{app: prelim} for preliminaries). 

Formally, given a multimodal input $x = (x_{\text{text}}, x_{\text{image}})$, we aim to optimize the MLLM policy network $\pi_\theta$ by maximizing a surrogate objective (e.g., the GRPO objective, detailed in Eq. \ref{eq: grpo},  Appendix \ref{app: prelim}).

The core of our approach is to enable targeted exploration in multimodal RLVR through dual-uncertainty guided policy learning. Rather than relying on passive data augmentation, DUPL explicitly explores where the model is uncertain. To this end, we transform visual perturbations into an active sensitivity probe that quantifies perceptual uncertainty. In parallel, we quantify output uncertainty. These two uncertainties serve as active feedback signals: they are integrated into the advantage function to guide policy optimization, incentivizing the model to explore perceptually and decision-wise ambiguous states. Finally, a dynamic branch prioritization schedule modulates this process, steering the model to prioritize exploration in early stages before stabilizing on the raw view for convergence. The full procedure is summarized in Algorithm~\ref{alg:dupl} in Appendix \ref{app: algo}.

% \vspace{-5pt}
\paragraph{Perceptual Uncertainty.}
% \vspace{-5pt}

% Moving beyond the standard use of image augmentation for RL \citep{yarats2021image, laskin2020reinforcement}, we introduce controlled perturbations to the image, and transform visual perturbations into an active sensitivity probe, which quantifies perceptual uncertainty by measuring the the extent to which the model’s output distribution varies under transformations of the image. Variation of the model’s predictions captures how sensitive the model’s predictions are to plausible visual perturbations, and therefore identify states worthy of exploration.

Moving beyond the standard use of image augmentation for RL~\citep{yarats2021image, laskin2020reinforcement}, we introduce controlled image perturbations and transform them into an active sensitivity probe that quantifies perceptual uncertainty by measuring how the model’s output distribution varies under visual transformations. Variations in the model’s predictions reflect its sensitivity to plausible perturbations, and therefore identify states worthy of exploration.

Specifically, for each image $x_{\text{image}}$ in the training dataset, we create a perturbed counterpart $x_{\text{image}}'$ through a stochastic augmentation function $\mathcal{T}$. This function applies a composition of transformations: $x_{\text{image}}' = \mathcal{T}(x_{\text{image}})$, where $\mathcal{T}$ includes random horizontal/vertical flips, rotations, color jittering, and the addition of Gaussian noise. Then we employ a dual-branch forward pass, as shown in Figure \ref{fig:app}. The raw branch processes the original input $x$ to produce an output probability distribution $p=\pi_\theta(\cdot|x)$, while the noisy branch uses the perturbed input $x'$ to produce a distribution $q=\pi_\theta(\cdot|x')$. After obtaining the two distributions, we represent the perceptual uncertainty $u_\text{per}$ as the divergence between them. We measure it using a symmetric KL divergence, which is calculated as the mean of the forward and backward KL divergences, encouraging exploration while maintaining stability:
\begin{equation} \label{eq: u_per}
    u_\text{per} = \frac{1}{2}\big(\text{D}_\text{KL}(p||q) + \text{D}_\text{KL}(q||p)\big).
\end{equation}

\paragraph{Output Uncertainty.}
After modeling the perceptual uncertainty, to promote general policy stochasticity and exploration in the action space, we model the output uncertainty $u_\text{out}$, which is based on the token entropy of the policy's output distribution: $u_\text{out} = - \sum_{v \in \mathcal{V}} \pi_{\theta} \left(v \mid x, o_{<t}\right)\log \pi_{\theta} \left(v \mid x, o_{<t}\right)$, where $\mathcal{V}$ denotes the vocabulary. 

% \vspace{-3pt}
\paragraph{Dual-Uncertainty Guided Policy Learning.} 

After obtaining the two uncertainties, we integrate them into the advantage function to guide policy learning. We maintain separate advantage calculations for the raw and noisy branches. For the raw branch, we compute the output guidance signal induced by the output uncertainty, which is defined as: 
$
    g_\text{out} = \min\!\Big(\frac{|A|}{\beta_o}, \alpha_o \cdot \mathrm{stopgrad}(u_\text{out}) \Big),
    \label{eq: g_out}
$
where $\alpha_o$ and $\beta_o$ are scaling factors, $\mathrm{stopgrad(\cdot)}$ is the stop gradient operator, modulating the update magnitude without affecting gradient propagation.

For the noisy branch, except the output guidance signal, a perceptual guidance signal is incorporated, derived from the measured perceptual uncertainty: 
$
    g_\text{per} = \min \! \Big(\frac{|A^\text{noi}|}{\beta_p}, \alpha_p \cdot \mathrm{stopgrad}(u_\text{per}) \Big),
    \label{eq: g_per}
$
where $A^\text{noi}$ is the advantage for the noisy branch, $\alpha_p$ and $\beta_p$ are scaling factors. 

Therefore, for the raw branch, the uncertainty-guided advantage is calculated as: 
$
    \hat{A}^\text{raw} = A^\text{raw} + g_\text{out}^\text{raw},
    \label{eq: adv_raw}
$
and for the noisy branch:
$
    \hat{A}^\text{noi} = A^\text{noi} + g_\text{out}^\text{noi} + g_\text{per}.
    \label{eq: adv_aug}
$ When implementing DUPL, we use GRPO's standard estimator to compute the base advantages $A^\text{noi}$ and $A^\text{raw}$ (see Eq.~\ref{eq:adv_norm} in Appendix \ref{app: prelim}). Policy updates are then performed using the corresponding uncertainty-guided advantage, depending on the branch selected at each training step.

\vspace{-3pt}
\paragraph{Dynamic Branch Prioritization.}
% \vspace{-5pt}
During training, it is crucial to balance the aggressive exploration driven by the noisy branch with the stable learning provided by the raw branch. A policy trained exclusively on the noisy branch may become overly stochastic and fail to converge, while a policy trained solely on the raw branch may not explore enough to find the optimal path. To manage this trade-off, we employ a dynamic branch prioritization strategy. At each training step, we stochastically choose which advantage estimate to use for the  policy update. We define $p_\text{noi}$ as the probability of prioritizing the noisy branch, which is expressed as: 
$
    p_{\text{noi}}(s) = \max\!\big(0,1-\tfrac{s}{s_{\text{total}}}\big),
$
where $s$ is the current training step, $s_\text{total}$ is the total training steps. This probability is decayed over the course of training. Initially, $p_\text{noi}$ is high to promote broad exploration of the state space. As training progresses, $p_\text{noi}$ is gradually decreased, causing the optimizer to favor the more stable advantage estimates from the raw branch. 

% This allows the policy to first explore and then fine-tune its reasoning based on the original, unperturbed data.

\vspace{-5pt}
\section{Experiments}
\vspace{-5pt}
\subsection{Experimental Setup} \label{exp: setup}
\vspace{-3pt}
\paragraph{Implementation Details.}
We conduct direct RL training on the Qwen2.5-VL-3B and 7B \citep{bai2025qwen25vltechnicalreport} models. The models are trained to generate responses in a structured format, where the reasoning process is enclosed within \(\texttt{<think></think>}\) tags and the final answer is presented in \texttt{\textbackslash boxed\{\}}. We train all models on the MMRL30k dataset \citep{zhu2025shuffle}, which contains around 30K samples. We  generate $5$ rollouts per input with a rollout batch size of $256$. The implementation builds on the framework EasyR1 \citep{zheng2025easyr1}.

To inject perturbation into images, we apply random horizontal/vertical flips, rotations, color jittering, and the addition of Gaussian noise with zero mean and standard deviation $\sigma=0.4$. A sensitivity analysis on different noise levels is provided in Section~\ref{sec: sen}. For more training details, please see Appendix \ref{app: train}.

\begin{table*}[t]
    \centering
    \resizebox{\textwidth}{!}{
    \begin{tabular}{lcccccccc}
        \toprule
        \textbf{Model} & \textbf{MathVerse} & \textbf{MathVista} & \textbf{WeMath} & \textbf{HalluBench} & \textbf{ChartQA} & \textbf{LogicVista} & \textbf{Avg.} \\
        \midrule
        \multicolumn{8}{c}{\textit{Open Source Models}} \\ 
        \midrule
        R1-Onevision-7B \citep{yang2025r1} & 46.0 & 63.9 & 61.8 & 67.2 & 78.3 & 45.5 & 60.5 \\
        % Vision-R1-7B \citep{huang2025vision} & 50.2  & 71.2 & - & 57.8 & 82.7 & 47.8 & - \\
        OpenVLThinker-7B \citep{deng2025openvlthinker} & 48.0 & 70.0 & 67.1 & 60.0 & 78.9 & 47.1 & 61.9 \\
        VLAA-Thinker-7B \citep{chen2025sft} & 48.2 & 68.3 & 67.7 & 70.0 & 80.2 & 47.3 & 63.6 \\
        % Shuffle-R1-Qwen-7B \citep{zhu2025shuffle} & 50.4 & 75.8 & 71.0 & 71.0 & 83.9 & 50.2 & 67.1 \\
        % VL-Rethinker-7B \citep{wang2025vl} & 49.5  & 73.3 & 57.8 & 69.5 & 81.0 & 48.4 & 63.3 \\
        MM-Eureka-Qwen-7B \citep{meng2025mm} & 50.3 & 71.2 & 65.6 & 66.4 & 79.9 & 47.3 & 63.5 \\
        % NoisyRollout-7B \citep{liu2025noisyrollout} & 51.0 & 72.6 & 69.6 & 70.1 & 81.5 & 47.5 & 65.4 \\
        PAPO-7B \citep{wang2025perception} & 52.0 & 73.7 & 67.6 & 71.0 & 79.6 & 47.1 & 65.2 \\
        % ThinkLite-VL-7B \citep{wang2025sota} & 47.3 & 71.9 & 69.2 & 70.9 & 81.4 &  48.5 &  64.9 \\
        \midrule
        \multicolumn{8}{c}{\textit{DUPL}} \\
        \midrule
        Qwen2.5-VL-3B \citep{bai2025qwen25vltechnicalreport} & 34.0 & 58.4 & 51.8 & 59.9 & 73.1 & 38.0 & 52.5 \\
        \quad + GRPO & 40.6  & 66.4 & 60.8 & 65.5 & 77.6 & 39.3 & 58.4 \\ 
        \quad + NoisyRollout & 41.8 & 67.7 & 62.0 & 66.1 & 77.7 & 42.3 & 59.6 \\
        \rowcolor{gray!15} \quad + DUPL (Ours) & \bf 43.3 & \bf 69.4 & \bf 64.1 & \bf 67.0 & \bf 78.4 & \bf 45.0 & \bf 61.2 \\
        \midrule
        Qwen2.5-VL-7B \citep{bai2025qwen25vltechnicalreport}& 45.8 & 67.2 & 63.2 & 65.2 & 79.8 & 45.5 & 61.1 \\
        \quad + GRPO & 48.0 & 70.6 & 68.5 & 68.6 & 81.5 & 46.0 &  63.9 \\ 
        \quad + NoisyRollout & 51.0 & 72.6 & 69.6 & 70.1 & 81.8 & 47.5 & 65.4 \\
        \rowcolor{gray!15} \quad + DUPL (Ours) & \bf 52.1 & \bf 74.2 & \bf 71.1 & \bf 71.0 & \bf 84.0 & \bf 48.7 & \bf 66.9 \\
        \bottomrule
    \end{tabular}
    }
    \vspace{-5pt}
        \caption{\textbf{Model accuracy on diverse visual mathematical and general-domain reasoning benchmarks.} Compared to the base models, DUPL improves accuracy by up to 12.3\% (Avg. 8.7\%) for the 3B model and up to 7.9\% (Avg. 5.8\%) for the 7B model across all evaluated tasks. Furthermore, DUPL consistently outperforms the strong baselines GRPO and NoisyRollout, with DUPL-7B achieving the best overall average performance.}
    \label{tab:main}
    \vspace{-12pt}
\end{table*}

\vspace{-5pt}
\paragraph{Evaluation.}
We evaluate performance across multiple multimodal reasoning benchmarks, including MathVerse \citep{zhang2024mathverse}, MathVista \citep{lu2023mathvista}, WeMath \citep{qiao2024we}, HallusionBench \citep{guan2024hallusionbench}, ChartQA \citep{masry2022chartqa}, LogicVista \citep{xiao2024logicvista}, ChartMuseum \citep{tang2026chartmuseum}, MMReason \citep{yao2025mmreason}, and VisuLogic \citep{xu2025visulogic}. We follow the evaluation protocol of \citet{zhu2025shuffle} and use Qwen2.5-72B-Instruct \citep{qwen2025qwen25technicalreport} to extract final answers from model responses and assess their correctness against reference answers.

We compare DUPL against three controlled baselines: (1) the off-the-shelf base models, (2) the strong RLVR baseline GRPO~\citep{shao2024deepseekmath}, and (3) NoisyRollout~\citep{liu2025noisyrollout}, a passive data augmentation method. We train NoisyRollout under identical experimental settings as DUPL. For broader context, we also include evaluation results from several external models: R1-Onevision-7B~\citep{yang2025r1}, OpenVLThinker-7B~\citep{deng2025openvlthinker}, VLAA-Thinker-7B~\citep{chen2025sft}, MM-Eureka-Qwen-7B~\citep{meng2025mm}, and PAPO-7B~\citep{wang2025perception}.

\vspace{-5pt}
\subsection{Main Results}
\vspace{-5pt}

% We first evaluate DUPL on multimodal mathematical reasoning benchmarks, including MathVerse, MathVista, and WeMath (Table~\ref{tab:main_1}). Compared to the Qwen2.5-VL base models, DUPL improves accuracy by up to 11.2\% (Avg. 10.1\%) for the 3B model and up to 7.9\% (Avg. 7.1\%) for the 7B model. DUPL also consistently outperforms the strong RLVR baseline GRPO across all benchmarks, indicating more effective policy learning for multimodal mathematical reasoning.

\begin{figure}[t]
    \centering
    % First subfigure
    \begin{subfigure}[b]{0.49\linewidth}
        \centering
        \includegraphics[width=\textwidth]{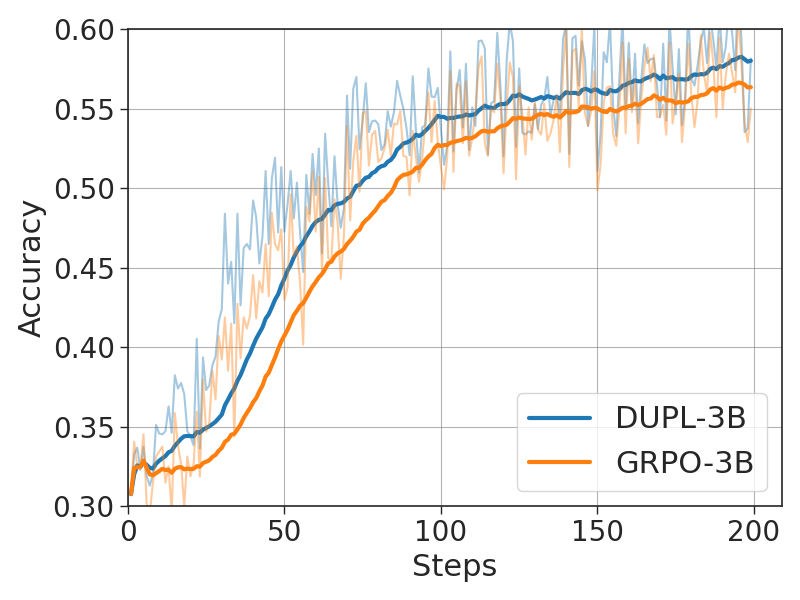}
        \caption{3B models}
        \label{fig:acc_3b}
    \end{subfigure}
    \hfill
    % Second subfigure
    \begin{subfigure}[b]{0.49\linewidth}
        \centering
        \includegraphics[width=\textwidth]{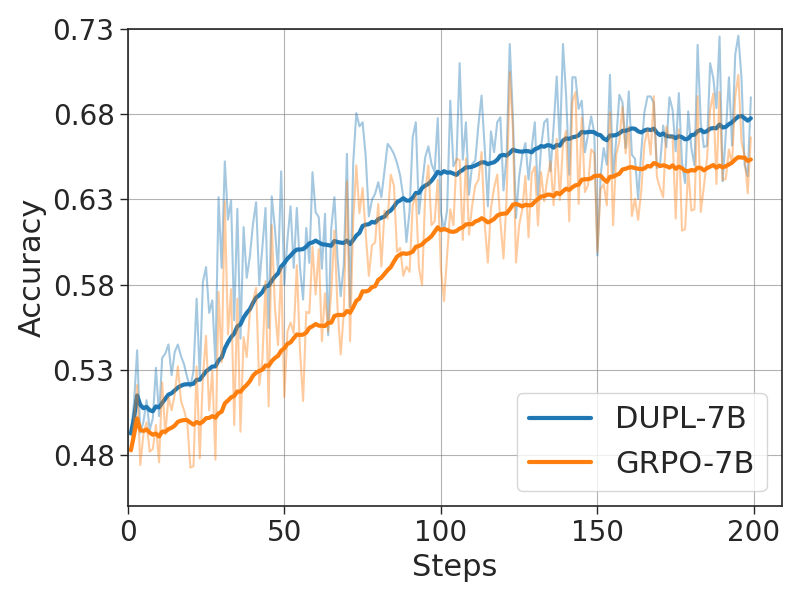}
        \caption{7B models}
        \label{fig:acc_7b}
    \end{subfigure}
    \caption{\textbf{Training accuracy of DUPL and GRPO on 3B and 7B models}. DUPL consistently achieves higher rewards than GRPO throughout training.}
    \label{fig:reward_acc}
\vspace{-15pt}
\end{figure}

We first evaluate DUPL across six benchmarks encompassing mathematical (MathVerse, MathVista, WeMath) and general-domain tasks (HallusionBench, ChartQA, LogicVista), as shown in Table~\ref{tab:main}. Compared to the base models, DUPL improves accuracy by up to 12.3\% (Avg. 8.7\%) for the 3B model and up to 7.9\% (Avg. 5.8\%) for the 7B model across all evaluated tasks. Furthermore, DUPL consistently outperforms GRPO and NoisyRollout, with DUPL-7B achieving the best overall average performance. Consistent with these results, the training accuracy curves in Figure~\ref{fig:reward_acc} show that DUPL maintains higher rewards than GRPO throughout training for both model scales, suggesting more effective policy learning. 

We also evaluate DUPL on three additional benchmarks: ChartMuseum \citep{tang2026chartmuseum}, MMReason \citep{yao2025mmreason}, and VisuLogic \citep{xu2025visulogic}. As detailed in Table~\ref{tab:additional_benchmarks} (Appendix~\ref{app: add}), DUPL improves the 7B base model by an average of 4.0\% across these benchmarks and outperforms the GRPO baseline. We report the training cost in Table~\ref{tab:cost} (Appendix~\ref{app: cost}), DUPL incurs only a modest computation overhead relative to GRPO, this minor trade-off is well justified by its large performance gains over the base models and its consistent outperformance of GRPO. Furthermore, we evaluate DUPL across six benchmarks using four random seeds (Table~\ref{tab:seed}, Appendix \ref{app: seed}). Results show that DUPL consistently enhances the 7B model and outperforms GRPO, showing its statistical significance.  

Additionally, we evaluate our framework on Qwen3-VL-Instruct 4B and 8B models. As shown in Table \ref{tab:qwen3} (Appendix \ref{app: qwen3}),  DUPL improves accuracy by up to 10.7\% (Avg.\ 5.3\%) for the 4B base model and up to 12.4\% (Avg.\ 6.3\%) for the 8B base model across all evaluated tasks, while consistently outperforming GRPO.

Taken together, these results demonstrate that DUPL yields consistent and robust improvements across both mathematical and general-domain multimodal reasoning tasks. The observed gains demonstrate that guiding exploration with perceptual and output uncertainty enables more effective policy learning.

% rather than relying on passive data augmentation or undirected exploration.

\vspace{-5pt}
\subsection{Ablation Studies} \label{sec: ablation}
\vspace{-5pt}
In this section, we conduct ablation studies to validate the contribution of each component in DUPL, including perceptual uncertainty, output uncertainty, and the dynamic branch prioritization strategy, as well as the impact of alternative divergence measures. Furthermore, to demonstrate the generalizability of our approach, we extend the experiments: we evaluate DUPL using an alternative RL algorithm, DAPO \citep{yu2026dapo}, and apply it to a different base model, LLaVA-OneVision-1.5-8B-Instruct \citep{an2025llava}.

\vspace{-4pt}
\paragraph{Perceptual Uncertainty.}

We first investigate the specific contribution of perceptual uncertainty by performing an ablation study where the term $u_{\text{per}}$ is deactivated. We report the quantitative results for mathematical reasoning and general-domain benchmarks in Table~\ref{tab:abl_math_ve} and Table~\ref{tab:abl_general_ve}, respectively. As shown, removing perceptual uncertainty from the feedback guidance results in an accuracy drop of up to 3.8\% (Avg. 1.7\%) on mathematical reasoning benchmarks. On general-domain reasoning tasks, performance decreases by up to 1.3\% (Avg. 1.0\%). We further illustrate the training dynamics in Figure~\ref{fig:acc_per} (Appendix \ref{app: abl}). Without perceptual uncertainty, the learning curve consistently lags behind that of the full DUPL approach, indicating slower and less effective policy optimization. These results confirm that incorporating perceptual uncertainty as a feedback signal plays a critical role in guiding policy learning. By explicitly steering exploration toward visually uncertain states, perceptual uncertainty enables more targeted exploration, thereby improving the model's reasoning accuracy.

\begin{table}[t]
    \centering

    % \vspace{-5pt}
    \resizebox{\linewidth}{!}{
    \begin{tabular}{lcccc}
        \toprule
        \textbf{Approach} & \textbf{MathVerse} & \textbf{MathVista} & \textbf{WeMath} & \textbf{Avg.}\\
        \midrule
        % GRPO & 48.0 & 72.1 & 69.5  &  63.2\\ 
        % \midrule
        % \multicolumn{5}{c}{\text{VOGUE}} \\
        Full approach & 52.1 & 74.2 & 71.1 & 65.8 \\
        \quad w/o $u_\text{per}$ &  48.3 & 73.6 & 70.3 & 64.1 \\
        \quad w/o $u_\text{out}$ & 48.6 & 73.5 & 70.8 & 64.3 \\
        \quad w/o $u_\text{per}$ \& $u_\text{out}$ & 48.0 & 73.1 & 68.5 & 63.2 \\
        \bottomrule
    \end{tabular}
    }
    \vspace{-5pt}
        \caption{\textbf{Ablations of DUPL accuracy on \emph{mathematical} reasoning benchmarks}. Removing perceptual uncertainty $u_\text{per}$ or output uncertainty $u_\text{out}$ reduces performance, with the largest drop when both are removed, confirming their complementary benefits.}
    \label{tab:abl_math_ve}
    \vspace{-5pt}
\end{table}

\begin{table}[t]
    \centering

    % \vspace{-5pt}
    \resizebox{\linewidth}{!}{
    \begin{tabular}{lcccc}
        \toprule
        \textbf{Approach} & \textbf{HalluBench} & \textbf{ChartQA} & \textbf{LogicVista} & \textbf{Avg.}\\
        \midrule
        % GRPO & 68.6 & 81.9 & 42.0  & 64.2 \\ 
        % \midrule
        % \multicolumn{5}{c}{\text{VOGUE}} \\
        Full approach & 71.0 & 84.0 & 48.7 & 67.9 \\
        \quad w/o $u_\text{per}$ &  69.7 & 83.4 & 47.8 & 66.9 \\
        \quad w/o $u_\text{out}$ & 70.2 & 82.4 & 47.8 & 66.8 \\
        \quad w/o $u_\text{per}$ \& $u_\text{out}$ & 69.2 & 82.1 & 46.4 & 65.9 \\
        \bottomrule
    \end{tabular}
    }
    \vspace{-5pt}
        \caption{\textbf{Ablations of DUPL on \emph{general-domain} reasoning benchmarks}. The same ablation trend in mathematical reasoning holds in general-domain settings, indicating that perceptual uncertainty and output uncertainty generalize beyond math tasks.}
    \label{tab:abl_general_ve}
    \vspace{-10pt}
\end{table}

\vspace{-5pt}
\paragraph{Output Uncertainty.}

Next, we evaluate the impact of output uncertainty by removing the term $u_{\text{out}}$. We report the results on visual mathematical reasoning benchmarks in Table~\ref{tab:abl_math_ve}, and the results on general-domain reasoning in Table~\ref{tab:abl_general_ve}. In the absence of output uncertainty guidance, accuracy drops by up to 3.5\% (Avg. 1.5\%) on mathematical benchmarks and by up to 1.6\% (Avg. 1.1\%) on general-domain tasks. As illustrated by the training accuracy curves in Figure~\ref{fig:acc_per} (Appendix \ref{app: abl}), the variant without output uncertainty lags behind compared to the full DUPL approach.

Furthermore, jointly removing both perceptual and output uncertainty results in a more pronounced performance degradation. As shown in Tables~\ref{tab:abl_math_ve} and~\ref{tab:abl_general_ve}, this combined ablation leads to accuracy drops of up to 4.1\% (Avg. 2.6\%) on mathematical reasoning benchmarks and up to 2.3\% (Avg. 2.0\%) on general-domain reasoning tasks.

These results demonstrate that both uncertainties provide effective and complementary feedback signals for policy optimization. Perceptual uncertainty encourages targeted exploration within the visual state space, while output uncertainty induces beneficial stochasticity in the textual output space.

\begin{table}[t]
    \centering

    % \vspace{-5pt}
    \resizebox{\linewidth}{!}{
    \begin{tabular}{lcccc}
        \toprule
        \textbf{Approach} & \textbf{MathVerse} & \textbf{MathVista} & \textbf{WeMath} & \textbf{Avg.}\\
        \midrule
        % GRPO & 48.0 & 72.1 & 69.5  &  63.2\\ 
        % \midrule
        % \multicolumn{5}{c}{\text{VOGUE}} \\
        Forward KL &  39.4 & 70.7 & 56.1 & 55.4 \\
        Symmetric KL & 52.1 & 74.2 & 71.1 & 65.8 \\
        \bottomrule
    \end{tabular}
    }
    \vspace{-5pt}
        \caption{\textbf{DUPL accuracy on \emph{mathematical} reasoning benchmarks under different divergence measures.} Symmetric KL provides stable uncertainty guidance and improves accuracy, while forward KL induces excessive divergence and degrades performance.}
    \label{tab:abl_math_kl}
    \vspace{-5pt}
\end{table}

\begin{table}[t]
    \centering

    % \vspace{-5pt}
    \resizebox{\linewidth}{!}{
    \begin{tabular}{lcccc}
        \toprule
        \textbf{Approach} & \textbf{HalluBench} & \textbf{ChartQA} & \textbf{LogicVista} & \textbf{Avg.}\\
        \midrule
        % GRPO & 68.6 & 81.9 & 42.0  & 64.2 \\ 
        % \midrule
        % \multicolumn{5}{c}{\text{VOGUE}} \\
        Forward KL &  67.5 & 80.3 & 45.1 & 64.3 \\
        Symmetric KL & 71.0 & 84.0 & 48.7 & 67.9 \\
        \bottomrule
    \end{tabular}
    }
    \vspace{-5pt}
        \caption{\textbf{DUPL accuracy on \emph{general-domain} reasoning benchmarks under different divergence measures.} Symmetric KL enables stable uncertainty-guided learning, while forward KL leads to excessive divergence and degraded performance.}
    \label{tab:abl_general_kl}
    \vspace{-12pt}
\end{table}

\begin{table*}[t]
\centering
\small
\resizebox{\linewidth}{!}{
\begin{tabular}{lccccccc}
\toprule
\textbf{Model} & \textbf{MathVerse} & \textbf{MathVista} & \textbf{WeMath} & \textbf{HalluBench} & \textbf{ChartQA} & \textbf{LogicVista} & \textbf{Avg.} \\
\midrule
Qwen2.5-VL-7B   & 45.8 & 67.2 & 63.2 & 65.2 & 79.8 & 45.5 & 61.1 \\
\quad + DAPO            & 50.0 & 74.1 & 69.5 & 69.8 & 82.2 & 48.9 & 65.8 \\
\quad + DUPL (w/ DAPO)  & \bf 52.8 & \bf 75.7 & \bf 69.6 & \bf 71.2 & \bf 84.1 & \bf 52.0 & \bf 67.6 \\
\bottomrule
\end{tabular}
}
\vspace{-5pt}
\caption{\textbf{Generalization to alternative RL algorithms.} When applied to the DAPO framework, DUPL improves the 7B base model by an average of 6.5\% across six benchmarks, consistently outperforming the DAPO baseline.}
\label{tab:dupl_dapo}
\vspace{-10pt}
\end{table*}

\begin{figure}[t]
    \centering
    % First subfigure
    \begin{subfigure}[b]{0.49\linewidth}
        \centering
        \includegraphics[width=\textwidth]{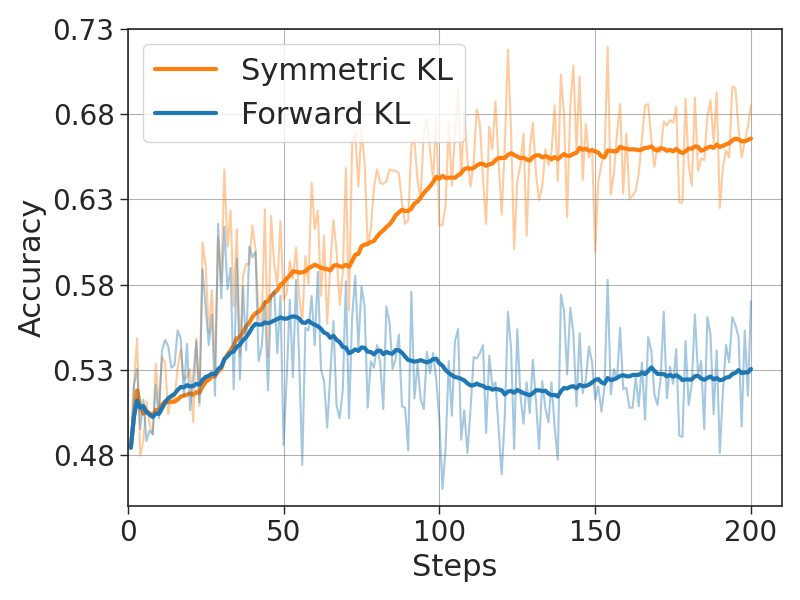}
        \caption{Training accuracy}
        \label{fig:kl_acc}
    \end{subfigure}
    \hfill
    % Second subfigure
    \begin{subfigure}[b]{0.49\linewidth}
        \centering
        \includegraphics[width=\textwidth]{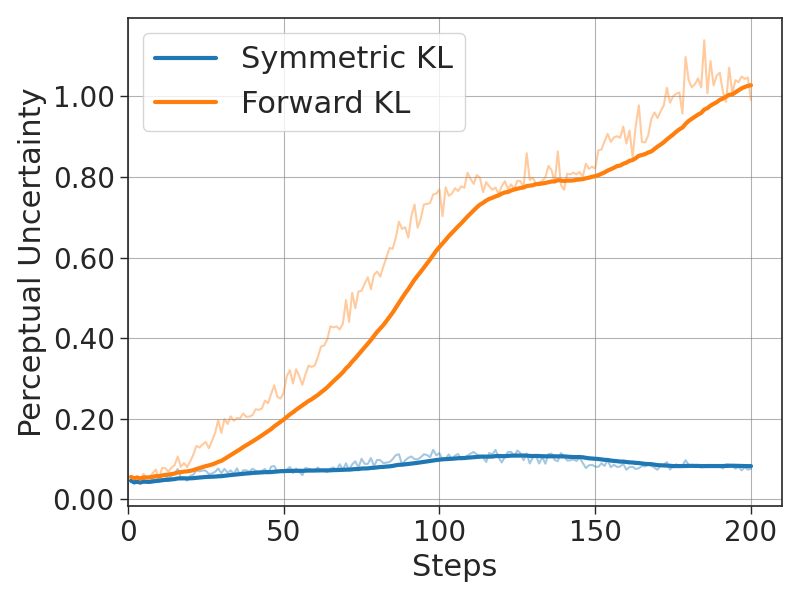}
        \caption{Perceptual uncertainty}
        \label{fig:forward_kl}
    \end{subfigure}
    % \vspace{-5pt}
    \caption{\textbf{Training accuracy and qualitative analysis of perceptual uncertainty under different divergence measures.} Measuring perceptual uncertainty with symmetric KL provides a stable signal that effectively guides policy learning while maintaining training stability. In contrast, forward KL produces excessively large perceptual uncertainty, resulting in unstable training and decreased accuracy.}
    \label{fig:kl_abl}
\vspace{-10pt}
\end{figure}

\vspace{-6pt}
\paragraph{Alternative Divergence Measures.}
To ablate the formulation of perceptual uncertainty, we replace the symmetric KL divergence with a forward KL variant. However, as shown in Figure \ref{fig:kl_acc}, the forward KL divergence leads to unstable training with accuracy declining. We qualitatively analyze the evolution of perceptual uncertainty under different KL formulations throughout training. As illustrated in Figure~\ref{fig:forward_kl}, perceptual uncertainty measured by forward KL grows excessively large, whereas uncertainty computed using symmetric KL exhibits a moderate increase followed by a gradual decrease, remaining overall stable during training. This observation explains the degraded performance under forward KL: it encourages the model to diverge excessively, leading to unstable policy updates. We also evaluate both variants on diverse reasoning benchmarks (Tables~\ref{tab:abl_math_kl} and~\ref{tab:abl_general_kl}). Consistent with Figure~\ref{fig:kl_abl}, the forward KL formulation results in lower accuracy.

Overall, this ablation validates the use of symmetric KL for measuring perceptual uncertainty, as it provides a stable uncertainty signal that supports effective exploration while preserving training stability.

\vspace{-8pt}
\paragraph{Dynamic Branch Prioritization.}

We evaluate the effect of the dynamic branch prioritization mechanism, which gradually adjusts the probability of selecting the noisy branch versus the raw branch. To isolate its effect, we replace it with a fixed sampling probability of $0.5$. We report the results on mathematical and general-domain reasoning benchmarks in Table~\ref{tab:abl_math_sample} and Table~\ref{tab:abl_general_sample}, respectively. As shown, dynamic branch prioritization consistently outperforms fixed-probability sampling, yielding accuracy improvements of up to 3.6\% (Avg. 2.5\%) on mathematical reasoning benchmarks and up to 1.8\% (Avg. 1.4\%) on general-domain reasoning benchmarks. Similar trends are observed in the training curves shown in Figure~\ref{fig:acc_branch} (Appendix~\ref{app: abl}).

These results highlight the advantage of dynamic branch prioritization: higher reliance on the noisy branch facilitates exploration during early training, while progressively emphasizing the raw branch stabilizes optimization and improves convergence in later stages.

\begin{table}[ht]
    \centering

    \vspace{-5pt}
    \resizebox{\linewidth}{!}{
    \begin{tabular}{lcccc}
        \toprule
        \textbf{Approach} & \textbf{MathVerse} & \textbf{MathVista} & \textbf{WeMath} & \textbf{Avg.}\\
        \midrule
        % GRPO & 48.0 & 72.1 & 69.5  &  63.2\\ 
        % \midrule
        % \multicolumn{5}{c}{\text{VOGUE}} \\
        Fixed Prob &  48.5 & 73.6 & 67.8 & 63.3 \\
        Branch Prioritization & 52.1 & 74.2 & 71.1 & 65.8 \\
        \bottomrule
    \end{tabular}
    }
    \vspace{-5pt}
        \caption{\textbf{DUPL accuracy on \emph{mathematical} reasoning benchmarks evaluating dynamic branch prioritization.} Dynamic branch prioritization outperforms fixed-probability sampling.}
    \label{tab:abl_math_sample}
    \vspace{-5pt}
\end{table}

\begin{table}[ht]
    \centering

    % \vspace{-5pt}
    \resizebox{\linewidth}{!}{
    \begin{tabular}{lcccc}
        \toprule
        \textbf{Approach} & \textbf{HalluBench} & \textbf{ChartQA} & \textbf{LogicVista} & \textbf{Avg.}\\
        \midrule
        Fixed Prob &  69.9 & 82.6 & 46.9 & 66.5 \\
        Branch Prioritization & 71.0 & 84.0 & 48.7 & 67.9 \\
        \bottomrule
    \end{tabular}
    }
    \vspace{-5pt}
        \caption{\textbf{DUPL accuracy on \emph{general-domain} reasoning benchmarks evaluating dynamic branch prioritization.} Dynamic branch prioritization outperforms fixed-probability sampling.}
    \label{tab:abl_general_sample}
    \vspace{-10pt}
\end{table}

\vspace{-6pt}
\paragraph{Alternative RL Algorithms.} 

To demonstrate that DUPL generalizes beyond GRPO, we evaluate it using DAPO as the underlying RL algorithm. As shown in Table \ref{tab:dupl_dapo}, integrating DUPL with DAPO improves the 7B base model by an average of 6.5\% across six benchmarks, consistently outperforming the standard DAPO baseline. These results highlight that our uncertainty-guided exploration is algorithm-agnostic and can effectively enhance other RLVR methods.

\begin{table*}[ht]
\centering
\resizebox{\linewidth}{!}{%
\begin{tabular}{lccccccc}
\toprule
\textbf{Model} & \textbf{MathVerse} & \textbf{MathVista} & \textbf{WeMath} & \textbf{HalluBench} & \textbf{ChartQA} & \textbf{LogicVista} & \textbf{Avg.} \\
\midrule
LLaVA-OneVision-1.5-8B & 45.6 & 68.8 & 61.6 & 65.7 & 81.6 & 46.0 & 61.6 \\
\quad + GRPO & 48.0 & 70.9 & 67.8 & 68.5 & 82.7 & 46.5 & 64.1 \\
\quad + DUPL & \bf 49.8 & \bf 73.5 & \bf 70.6 & \bf 70.8 & \bf 84.8 & \bf 48.2 & \bf 66.3 \\
\bottomrule
\end{tabular}%
}
\vspace{-5pt}
\caption{\textbf{Evaluation performance using LLaVA-OneVision-1.5-8B.} DUPL remains effective with this architecture, improving base model accuracy by an average of 4.7\% and consistently outperforming GRPO across all tasks.}
\label{tab:llava}
\vspace{-5pt}
\end{table*}

\vspace{-5pt}
\paragraph{Alternative Base Models.}

To evaluate the generalizability of DUPL, we evaluate it on an alternative base model, LLaVA-OneVision-1.5-8B-Instruct \citep{an2025llava}. As shown in Table \ref{tab:llava}, our method remains effective, improving the base model by an average of 4.7\% across the six benchmarks. Furthermore, DUPL consistently outperforms the GRPO baseline, verifying its broad applicability across diverse architectures.

\begin{table*}[t]
    \centering
    \resizebox{0.8\textwidth}{!}{
    \begin{tabular}{lccccccc}
        \toprule
        \textbf{Approach} & \textbf{MathVerse} & \textbf{MathVista} & \textbf{WeMath} & \textbf{HalluBench} & \textbf{ChartQA} & \textbf{LogicVista} & \textbf{Avg.} \\
        \midrule
        $\sigma=0.2$ & 48.4 & 74.0 & 68.8 & 69.4 & 81.9 & 45.3 & 64.6 \\
        $\sigma=0.4$ & \bf 52.1 & \bf 74.2 & \bf 71.1 & \bf 71.0 & \bf 84.0 & \bf 48.7 & \bf 66.9 \\
        $\sigma=0.8$ & 49.2 & 73.5 & 66.8 & 70.4 & 82.9 & 46.2 & 64.8 \\
        \bottomrule
    \end{tabular}
    }
    \vspace{-5pt}
    \caption{\textbf{DUPL accuracy on diverse reasoning benchmarks with different noise levels.} Moderate noise ($\sigma=0.4$) yields the best accuracy across all evaluated benchmarks.}
    \label{tab:abl_noise_merged}
    \vspace{-13pt}
\end{table*}

\vspace{-6pt}
\subsection{Sensitivity Analysis} \label{sec: sen}
\vspace{-6pt}

\begin{figure}[ht]
    \centering
    % First subfigure
    \begin{subfigure}[b]{0.49\linewidth}
        \centering
        \includegraphics[width=\textwidth]{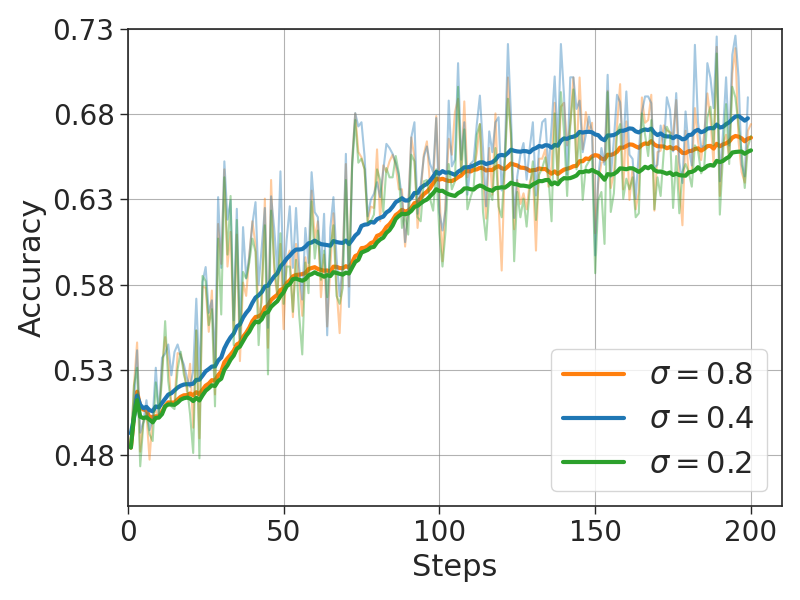}
        \caption{Training accuracy}
        \label{fig:acc_noise}
    \end{subfigure}
    \hfill
    % Second subfigure
    \begin{subfigure}[b]{0.49\linewidth}
        \centering
        \includegraphics[width=\textwidth]{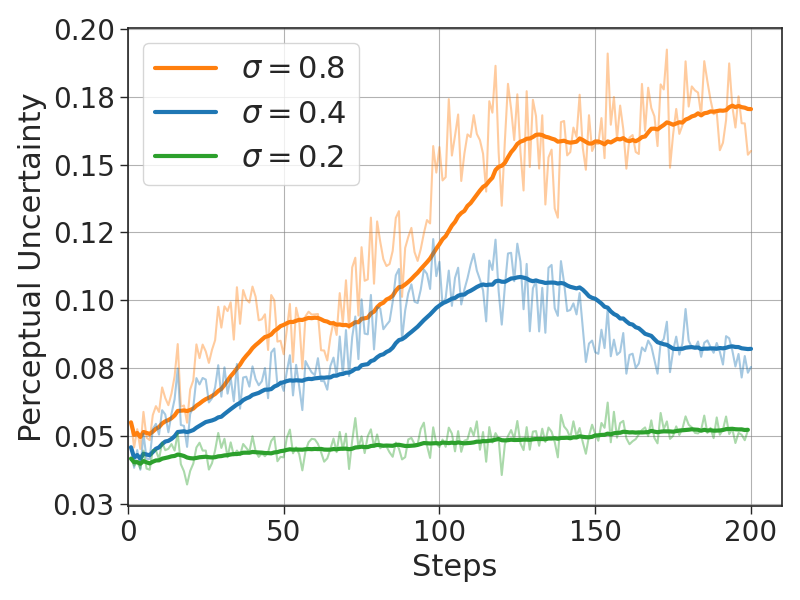}
        \caption{Perceptual uncertainty}
        \label{fig:upper_noise}
    \end{subfigure}
    \caption{\textbf{Training accuracy and perceptual uncertainty across noise levels}. Moderate noise ($\sigma=0.4$) yields the best performance. Low noise ($\sigma=0.2$) provides insufficient visual exploration, whereas high noise ($\sigma=0.8$) introduces excessive variance that overly amplifies perceptual uncertainty.}
    \label{fig:noise_abl}
\vspace{-13pt}
\end{figure}

We conduct a sensitivity analysis of DUPL under varying levels of noise by adjusting the standard deviation of the Gaussian perturbation in the noisy branch ($\sigma \in \{0.2, 0.4, 0.8\}$). As reported in Table~\ref{tab:abl_noise_merged}, moderate noise ($\sigma=0.4$) achieves the highest accuracy. Low noise ($\sigma=0.2$) provides insufficient visual exploration, whereas high noise ($\sigma=0.8$) introduces excessive variance that overly amplifies perceptual uncertainty. This trend is further supported by the training accuracy curves in Figure~\ref{fig:acc_noise}, as well as the qualitative analysis of perceptual uncertainty evolution over training under different noise levels in Figure~\ref{fig:upper_noise}. 

\vspace{-5pt}
\section{Related Work}
\vspace{-5pt}

\paragraph{Exploration in Text-Based Reasoning.}
Recent text-only RLVR works have begun addressing policy exploration. To complement outcome rewards, i-MENTOR~\citep{gao2025navigate} uses trajectory-aware intrinsic signals and dynamic reward scaling, while Retrospective Replay~\citep{dou2025improving} revisits promising early states to counter exploration decay. Other exploration strategies leverage outcome-based schemes~\citep{song2025outcome} or granular process rewards~\citep{setlur2024rewarding}, though scoring intermediate steps remains challenging. Additionally, upweighting negative samples can mitigate diversity collapse~\citep{zhu2025surprising}, and EVOL-RL promotes reasoning diversity through novelty-oriented reinforcement without explicit labels~\citep{zhou2025evolving}.

\vspace{-6pt}
\paragraph{Multimodal RLVR.}
RLVR is increasingly applied to enhance multimodal reasoning. \citet{yang2025r1} and \citet{huang2025vision} extended language reasoning with visual inputs and vision-grounded prompts, while others focused on cross-modal generalization \citep{deng2025openvlthinker} and unifying visual-textual signals \citep{chen2025sft}. Further advances include self-rewarding mechanisms that decompose visual and linguistic reasoning \citep{li2025self}, perception-aware losses for better visual grounding \citep{wang2025perception}, and single-rollout RL for improved training efficiency \citep{liu2025stable}.

Despite these advances, effective exploration in multimodal RLVR remains largely underexplored. Existing methods generally treat visual inputs as deterministic, overlooking inherent perceptual ambiguity. While recent works inject noise into training images to encourage exploration \citep{liu2025noisyrollout, yao2025r1}, they rely on passive data augmentation. The policy learning objective remains unchanged, exploration is global and undirected.

In contrast, DUPL introduces targeted exploration for multimodal RLVR via dual-uncertainty guided policy learning. By jointly modeling perceptual uncertainty and output uncertainty and incorporating them into an uncertainty-driven feedback loop, DUPL actively regulates policy updates and directs exploration toward states where the model is uncertain, addressing a key limitation of prior multimodal RLVR methods.

% While some studies identify insufficient exploration in RL algorithms such as GRPO \citep{shao2024deepseekmath} and propose dynamic KL regularization \citep{liu2025othink} or rule-based process rewards \citep{zhang2025r1}, these mechanisms primarily stabilize optimization rather than explicitly encourage exploration.

% coupling exploration to quantified visual uncertainty and, to our knowledge, is among the first modality-aware exploration frameworks for RLVR. Moreover, VOGUE is complementary to other language-side exploration strategies (e.g., temperature scheduling~\citep{liao2025enhancing}, KL regularization~\citep{liu2025scaling}, and output-level diversity/novelty bonuses~\citep{li2025jointly}, which have shown benefits mainly in text RL and can be combined for further gains.

% \vspace{-5pt}
\vspace{-5pt}
\section{Conclusions}
\vspace{-5pt}

% In conclusion, we introduce DUPL, a dual-uncertainty guided policy learning approach for multimodal RLVR. By measuring and leveraging both perceptual and output uncertainty, DUPL establishes an uncertainty-driven feedback loop that directs exploration toward states with genuine ambiguity, moving beyond passive data augmentation. Our approach integrates a dynamic branch prioritization mechanism and advantage shaping to effectively guide policy updates. Extensive experiments on six multimodal mathematical and general-domain reasoning benchmarks show that DUPL largely enhances the Qwen2.5-VL 3B and 7B models, achieving accuracy gains of up to 11.2\% on visual math tasks and up to 7.1\% on general-domain reasoning tasks, while consistently outperforming strong baselines. These results demonstrate the effectiveness of DUPL.

In conclusion, we introduce DUPL, a dual-uncertainty guided policy learning approach that overcomes the limitations of undirected exploration in multimodal RLVR. By measuring both perceptual and output uncertainty, DUPL creates a targeted feedback loop, utilizing dynamic branch prioritization and advantage shaping to guide policy updates. Extensive evaluation across diverse mathematical and general-domain benchmarks demonstrate that DUPL yields solid improvements. It achieves accuracy gains of up to 12.3\% (3B) and 7.9\% (7B) on Qwen2.5-VL, alongside up to 10.7\% (4B) and 12.4\% (8B) on Qwen3-VL-Instruct. Validating its broad applicability, we further demonstrate that DUPL successfully enhances an alternative base model, LLaVA-OneVision-1.5-8B, by 4.7\% on average, and effectively generalizes to the DAPO algorithm with an average gain of 6.5\%. These results establish DUPL as an effective approach of targeted exploration for multimodal RLVR.

\section*{Limitations}
While DUPL effectively demonstrates the value of dual-uncertainty guided policy learning, it presents several promising avenues for future exploration. Consistent with standard practice in prior work \citep{yarats2021image, laskin2020reinforcement, liu2025noisyrollout}, our current implementation employs augmentation techniques such as geometric manipulations and Gaussian noise, to induce perceptual uncertainty. A promising avenue for future investigation lies in exploring more sophisticated perturbation strategies or learnable noise generators that could expose subtler reasoning vulnerabilities. Furthermore, extending the dual-uncertainty framework to dynamic modalities, such as video, presents a worthy direction for establishing a more  comprehensive future work.
\bibliography{ref}

% while we successfully unify perceptual and output uncertainty, our framework currently treats the textual input as fixed; extending this dual-uncertainty paradigm to incorporate textual input perturbations or

%%%%%%%%%%%%%%%%%%%%%%%%%%%%%%%%%%%%%%%%%%%%%%%%%%%%%%%%%%%%%%%%%%%%%%%%%%%%%%%
%%%%%%%%%%%%%%%%%%%%%%%%%%%%%%%%%%%%%%%%%%%%%%%%%%%%%%%%%%%%%%%%%%%%%%%%%%%%%%%
% APPENDIX
%%%%%%%%%%%%%%%%%%%%%%%%%%%%%%%%%%%%%%%%%%%%%%%%%%%%%%%%%%%%%%%%%%%%%%%%%%%%%%%
%%%%%%%%%%%%%%%%%%%%%%%%%%%%%%%%%%%%%%%%%%%%%%%%%%%%%%%%%%%%%%%%%%%%%%%%%%%%%%%

\clearpage
\appendix

\section{Appendix}

\subsection{GRPO Preliminaries} \label{app: prelim}

In GRPO, given an input $x$, a group of responses $\{o_i\}_{i=1}^G$ are sampled from the old policy $\pi_{\theta_\text{old}}$, each associated with a reward $r_i$. Then the normalized advantage for response $o_i$ is defined as:
\vspace{-5pt}
\begin{align} \label{eq:adv_norm}
    A_i = \frac{r_i - \text{mean}(\{r_i\}_{i=1}^G)}{\text{std}(\{r_i\}_{i=1}^G)}.
\end{align}  
\vspace{-5pt}
As in PPO, GRPO uses clipped importance sampling to stabilize policy updates.  
Let $\rho_i(\theta) = \frac{\pi_\theta(o_i \mid x)}{\pi_{\theta_{\text{old}}}(o_i \mid x)}$ denote the probability ratio between the new and old policies. The GRPO objective is to maximize the following equation: 

% \begin{align} \label{eq: grpo}
% \mathcal{J}_{\text{GRPO}}(\theta) 
% = \mathbb{E}_{x \sim \mathcal{D}, \{o_i\} \sim \pi_{\theta_\text{old}}(\cdot \mid x)}
%     \left[
%         \frac{1}{G} \sum_{i=1}^G 
%         \min\!\left(
%             \rho_i(\theta) A_i,\; \\ 
%             \text{clip}\big(\rho_i(\theta), 1 - \epsilon_{\text{clip}}, 1 + \epsilon_{\text{clip}}\big) A_i
%         \right)
%     \right],
% \end{align}

\begin{align} \label{eq: grpo}
\mathcal{J}_{\text{GRPO}}(\theta) = & \mathbb{E}_{x \sim \mathcal{D}, \{o_i\} \sim \pi_{\theta_{\text{old}}}} \bigg[ \frac{1}{G} \sum_{i=1}^G \min \Big( \rho_i(\theta) A_i, \nonumber \\
& \text{clip}(\rho_i(\theta), 1 - \epsilon, 1 + \epsilon) A_i \Big) \bigg],
\end{align}

where $\epsilon_{\text{clip}}$ is the clipping hyperparameter.

% \newpage 
\subsection{Algorithm} \label{app: algo}
We present the full procedure of DUPL in Algorithm~\ref{alg:dupl}, summarizing Section~\ref{sec: app}. First, we measure output and perceptual uncertainty, which are then used to compute guidance signals. These signals are employed to shape the advantage for the raw and noisy branches separately. A dynamic branch prioritization mechanism is then applied to select which uncertainty-guided advantage is used for policy updates at each training step.

\subsection{Analysis} \label{app: analysis}

We express the policy gradient for the noisy branch as follows: 
\begin{equation} \label{eq:grad_dupl}
\begin{aligned}
    \nabla_{\theta} \mathcal{J}_{\text{DUPL}}(\theta) \propto & \ \mathbb{E}_{o' \sim \pi_{\theta_{\text{old}}}} \Big[ (A^{\text{noi}} + g_\text{out}^{\text{noi}} + g_\text{per}) \\
    & \cdot \nabla_{\theta} \log \pi_{\theta}(o'\mid x') \Big].
\end{aligned}
\end{equation}

From a gradient perspective, this formulation encourages more effective policy updates for targeted exploration compared to standard RLVR approaches such as GRPO. We omit caps/clipping for clarity. The term $g_{\text{per}}$ explicitly reweights the gradient to favor trajectories originating from states with high perceptual ambiguity, guiding the model to acquire more informative visual features. The term $g^{\text{noi}}_{\text{out}}$ acts as a general-purpose exploration mechanism in the token space: by rewarding higher entropy in the output distribution, it prevents premature convergence and complements the visual exploration induced by $g_{\text{per}}$. Unlike GRPO, which rely solely on the extrinsic reward signal $A^{\text{noi}}$, this decomposition shows that DUPL optimizes a composite objective that enables targeted exploration.

% The term $g_\text{per} \nabla_{\theta} \log \pi_{\theta}(o'\mid x')$ explicitly encourages the policy to increase the probability of action sequences that follow from perceptually uncertain states, guiding the model to acquire more informative visual features. There is also an term $g_\text{out}$. This acts as a general-purpose exploration mechanism in the action space that complements the exploration driven by $g_\text{per}$. While $g_\text{per}$ directs exploration toward perceptual uncertainty, $g_\text{out}$ maintains stochasticity in the textual output space.

\begin{algorithm*}[ht]
\caption{Dual-Uncertainty Guided Policy Learning (DUPL)}
\label{alg:dupl}
\begin{algorithmic}[1]
\Require Dataset $\mathcal{D}$, group size $G$, total training steps $s_{\text{total}}$, augmentation function $\mathcal{T}$.
\State Initialize policy parameters $\theta$, old policy $\theta_{\text{old}} \gets \theta$
\For{$s = 1$ to $s_{\text{total}}$}
    \State Sample input $x=(x_{\text{text}}, x_{\text{image}}) \sim \mathcal{D}$
    \State Construct perturbed image $x'_{\text{image}} = \mathcal{T}(x_{\text{image}})$, $x'=(x_{\text{text}},x'_{\text{image}})$
    \State Sample group of responses $\{o_i\}_{i=1}^G \sim \pi_{\theta_{\text{old}}}(\cdot \mid x)$, $\{o_i'\}_{i=1}^G \sim \pi_{\theta_{\text{old}}}(\cdot \mid x')$
    \State Compute branch prioritization probability:
        $
            p_{\text{noi}}(s) = \max\!\big(0,1-\tfrac{s}{s_{\text{total}}}\big)
        $
    % \State Compute rewards $\{r_i\}_{i=1}^G$ and group-normalized advantages $\{A_i\}_{i=1}^G$ (Eq.~\ref{eq:adv_norm})
    \For{$i=1$ to $G$}
        \State Compute raw advantage $A_i^{\text{raw}}$, noisy advantage $A_i^{\text{noi}}$ (Eq.~\ref{eq:adv_norm})
        \State Compute output uncertainty $u_{\text{out},i}^{\text{raw}}$ and $u_{\text{out},i}^{\text{noi}}$ for each branch
        \State Compute perceptual uncertainty $u_{\text{per},i}$ using token-level symmetric KL divergence (Eq.~\ref{eq: u_per})
        \State Compute guidance signals:
        \vspace{-10pt}
        \begin{align*}
            g_{\text{per},i} &= \min\!\Big(\tfrac{|A_i^{\text{noi}}|}{\beta_p}, \alpha_p \cdot \mathrm{stopgrad}(u_{\text{per},i})\Big)
        \end{align*}
        \vspace{-20pt}
        \begin{align*}
            g_{\text{out},i}^{\text{raw}} &= \min\!\Big(\tfrac{|A_i^{\text{raw}}|}{\beta_o}, \alpha_o \cdot \mathrm{stopgrad}(u_{\text{out},i}^{\text{raw}})\Big),
            g_{\text{out},i}^{\text{noi}} = \min\!\Big(\tfrac{|A_i^{\text{noi}}|}{\beta_o}, \alpha_o \cdot \mathrm{stopgrad}(u_{\text{out},i}^{\text{noi}})\Big)
        \end{align*}
        \vspace{-10pt}
        \State Compute shaped advantages:
        \begin{align*}
            \widehat{A}_i^{\text{raw}} = A_i^{\text{raw}} + g_{\text{out},i}^{\text{raw}}, \quad 
            \widehat{A}_i^{\text{noi}} = A_i^{\text{noi}} + g_{\text{out},i}^{\text{noi}} + g_{\text{per},i}
        \end{align*}
        \State Sample branch selector $z_i \sim \text{Bernoulli}(p_{\text{noi}}(s))$
        \State Select final advantage (with corresponding selected ($x$, $o_i$) or ($x'$, $o'_i$)): 
        \[
            \widehat{A}_i = z_i \cdot \widehat{A}_i^{\text{noi}} + (1-z_i)\cdot \widehat{A}_i^{\text{raw}}
        \]
    \EndFor
    \State Compute surrogate objective with shaped advantages (Eq.~\ref{eq: grpo})
    \State Update policy parameters $\theta \gets \theta + \eta \nabla_\theta \mathcal{J}_\text{DUPL}(\theta)$
    % \State Periodically update old policy $\theta_{\text{old}} \gets \theta$
\EndFor
\end{algorithmic}
\end{algorithm*}
 
\subsection{Prompt Templates} \label{app: prompt}
We list below the prompt used to instruct the model to produce the structured outputs.

\begin{tcolorbox}[colback=gray!10, colframe=gray!70, title=System Prompt]
You FIRST think about the reasoning process as an internal monologue and then provide the final answer. The reasoning process MUST be enclosed within \(\texttt{<think></think>}\) tags. The final answer MUST be put in \texttt{\textbackslash boxed\{\}}.
\end{tcolorbox}

\begin{figure}[ht]
    \centering
    \includegraphics[width=0.8\linewidth]{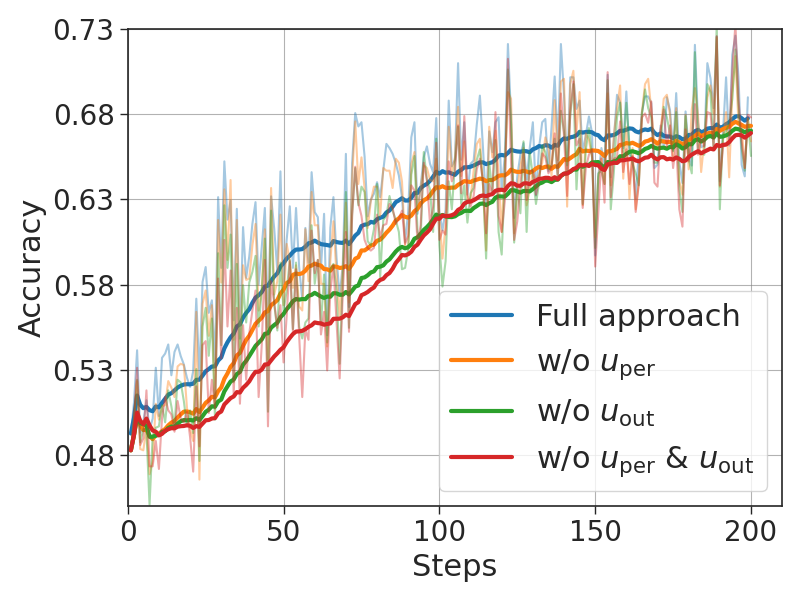}
    \caption{\textbf{Training dynamics of accuracy rewards under perceptual uncertainty $u_{\text{per}}$ and output uncertainty $u_{\text{out}}$.} Each uncertainty signal provides effective feedback guidance and improves performance, while their combination of full approach yields the best results.}
    \label{fig:acc_per}
\end{figure}

\begin{figure}[ht]
    \centering
    \includegraphics[width=0.8\linewidth]{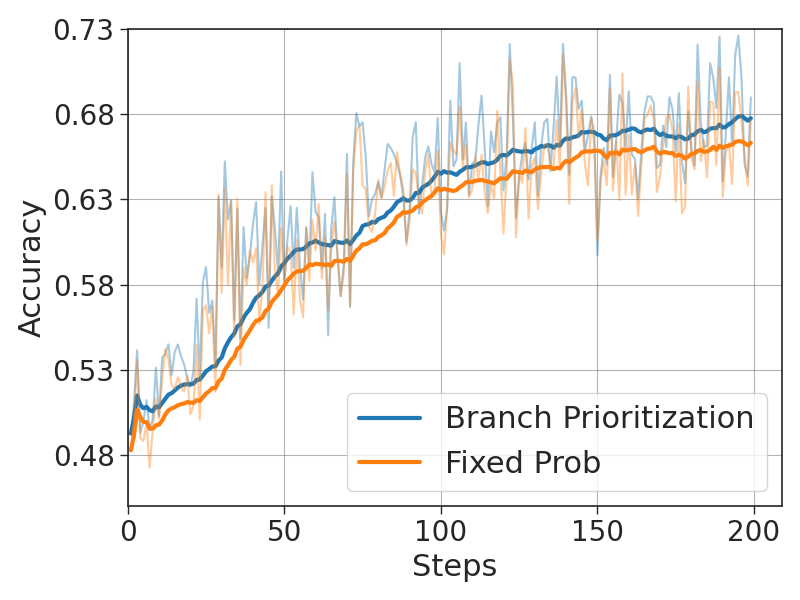}
    \caption{\textbf{Training dynamics of accuracy rewards comparing dynamic branch prioritization with fixed-probability sampling.} Dynamic branch prioritization consistently achieves higher accuracy.}
    \label{fig:acc_branch}
\end{figure}

\subsection{Training Details} \label{app: train}
% \vspace{-5pt}
We train all models on the $\text{MMRL30k}$ dataset \citep{zhu2025shuffle}, which contains around 30K samples. The models are trained to generate responses in a structured format, where the reasoning process is enclosed within \(\texttt{<think></think>}\) tags and the final answer is presented in \texttt{\textbackslash boxed\{\}}. The reward function for RL training is a combination of a format reward and an accuracy reward, with coefficients of \(0.1\) and \(0.9\), respectively. The training is performed for $200$ steps using the AdamW optimizer \citep{loshchilov2018decoupled} with a learning rate of $1e-6$ and a weight decay of $0.01$. We adopt a global batch size of $128$, a rollout batch size of $256$, and generate $5$ rollouts per input with a rollout temperature $1.0$. The implementation builds on the framework EasyR1 \citep{zheng2025easyr1}. 

When transferring perceptual and output uncertainty into guidance signals for advantage shaping to guide policy learning, we use $\alpha_p=1.0$ and $\beta_p=2.0$ for perceptual uncertainty, and $\alpha_o=0.4$ and $\beta_o=2.0$ for output uncertainty. A detailed sensitivity analysis of these parameters is provided in Appendix~\ref{app: sen}.

\subsection{Additional Sensitivity Analysis} \label{app: sen}
% To investigate the impact of the uncertainty feedback guidance signals parameters for advantage-shaping on the performance of DUPL, we conduct a systematic grid search of sensitivity analysis using the Qwen2.5-VL-7B model. Our framework introduces two sets of scaling factors: $(\alpha_o, \beta_o)$ which regulate the token-level output entropy guidance $g_\text{out}$, and $(\alpha_p, \beta_p)$ which scale the symmetric KL divergence guidance $g_\text{per}$ for capturing perceptual uncertainty.  

% We first analyze the performance of DUPL under different output uncertainty parameters. We vary $\alpha_o \in \{0.2, 0.4, 0.8\}$ and $\beta_o \in \{1.0, 2.0, 4.0\}$. As shown in Table \ref{tab:sensitivity_output}, moderate values ($\alpha_o=0.4, \beta_o=2.0$) yield the best performance. Then we analyze the performance of DUPL under different perceptual uncertainty parameters. We vary $\alpha_p \in \{0.5, 1.0, 2.0\}$ and $\beta_p \in \{1.0, 2.0, 4.0\}$. As shown in Table \ref{tab:sensitivity_perceptual}, $\alpha_p=1.0 and \beta_p=2.0$ yield the best performance.

To investigate the impact of the uncertainty feedback guidance parameters on advantage shaping, we conduct a systematic sensitivity analysis via grid search using the Qwen2.5-VL-7B model. Our framework introduces two sets of scaling factors: $(\alpha_o, \beta_o)$, which regulate the token-level output entropy guidance $g_\text{out}$, and $(\alpha_p, \beta_p)$, which modulate the symmetric KL divergence guidance $g_\text{per}$ for capturing perceptual uncertainty. 

We first analyze the effect of the output uncertainty parameters by varying the scaling factor $\alpha_o \in \{0.2, 0.4, 0.8\}$ and $\beta_o \in \{1.0, 2.0, 4.0\}$. As shown in Table~\ref{tab:sensitivity_output}, moderate values ($\alpha_o=0.4, \beta_o=2.0$) yield the best performance. Setting $\alpha_o$  low restricts action-space stochasticity, whereas high values introduce noise into the advantage estimation, both of which degrade final reasoning accuracy. Next, we analyze the performance of DUPL under varying perceptual uncertainty parameters by sweeping the coefficients $\alpha_p \in \{0.5, 1.0, 2.0\}$ and $\beta_p \in \{1.0, 2.0, 4.0\}$. As demonstrated in Table~\ref{tab:sensitivity_perceptual}, the configuration of $\alpha_p=1.0$ and $\beta_p=2.0$ achieves the highest accuracy. 

% This indicates that while targeted exploration toward states with high visual ambiguity is beneficial, over-scaling the perceptual guidance signal can destabilize policy optimization during the rollout updates.

\begin{table}[ht]
\centering
\small
\resizebox{\linewidth}{!}{
\begin{tabular}{cccccc}
\toprule
$\alpha_o$ & $\beta_o$ & \bf MathVerse & \bf MathVista & \bf ChartQA & \bf Avg. \\ \midrule
0.2        & 2.0       & 50.1      & 72.4      & 82.1    & 68.2    \\
\textbf{0.4} & \textbf{2.0} & \textbf{52.1} & \textbf{74.2} & \textbf{84.0} & \textbf{70.1} \\
0.8        & 2.0       & 48.9      & 71.5      & 81.3    & 67.2    \\
0.4        & 1.0       & 51.8      & 73.0      & 83.1    & 69.3    \\
0.4        & 4.0       & 51.2      & 73.5      & 83.4    & 69.4    \\ \bottomrule
\end{tabular}
}
\caption{\textbf{Sensitivity analysis of output uncertainty parameters ($\alpha_o, \beta_o$)}. $\alpha_o = 0.4$ and $\beta_o = 2.0$ yield the best performance.}
\label{tab:sensitivity_output}
\end{table}

\begin{table}[ht]
\centering
\small
\resizebox{\linewidth}{!}{
\begin{tabular}{cccccc}
\toprule
$\alpha_p$ & $\beta_p$ & \bf MathVerse & \bf MathVista & \bf ChartQA & \bf Avg. \\ \midrule
0.5        & 2.0       & 50.5      & 72.9      & 81.8    & 68.4    \\
\textbf{1.0} & \textbf{2.0} & \textbf{52.1} & \textbf{74.2} & \textbf{84.0} & \textbf{70.1} \\
2.0        & 2.0       & 50.4      & 72.2      & 81.9    & 68.2    \\
1.0        & 1.0       & 51.0      & 73.2      & 82.9    & 69.0    \\
1.0        & 4.0       & 51.5      & 73.6      & 83.5    & 69.5    \\ \bottomrule
\end{tabular}
}
\caption{\textbf{Sensitivity analysis of perceptual uncertainty parameters ($\alpha_p, \beta_p$)}. $\alpha_p = 1.0$ and $\beta_p = 2.0$ yield the best performance.}
\label{tab:sensitivity_perceptual}
\end{table}

\subsection{Training Cost Analysis} \label{app: cost}

We compare the training cost of DUPL and GRPO by measuring training time (minutes per step) and throughput (tokens processed per second per GPU). As shown in Table \ref{tab:cost}, DUPL incurs only a modest computational overhead compared to GRPO, reflected in a slight increase in training time per step and a corresponding reduction in throughput. However, this minor overhead is well justified by the large performance gains DUPL provides over the base models and its consistent outperformance of the GRPO baseline. 

\begin{table}[h]
\centering
\resizebox{\linewidth}{!}{%
\begin{tabular}{lcc}
\toprule
Method & Time (mins/step) & Throughput (tokens/s/GPU) \\
\midrule
GRPO & 4.12 & 496.14 \\
DUPL & 4.95 & 453.70 \\
\bottomrule
\end{tabular}
}
\caption{\textbf{Training cost comparison.} DUPL incurs only a modest computational overhead relative to GRPO, resulting in slightly increased training time per step and a corresponding reduction in throughput. This minor trade-off is well justified by the large performance gains DUPL achieves over the base models and its consistent outperformance of the GRPO baseline.}
\label{tab:cost}
\end{table}

\vspace{-5pt}
\subsection{Ablation Studies} \label{app: abl}
In Section \ref{sec: ablation}, we conduct a comprehensive set of ablation studies to validate the contribution of each component in DUPL, including perceptual uncertainty, output uncertainty, the dynamic branch prioritization strategy, and the choice of divergence measures. Here, we present additional results that further validate the role of some components.

In Figure \ref{fig:acc_per}, we show the training dynamics of accuracy under perceptual uncertainty $u_{\text{per}}$ and output uncertainty $u_{\text{out}}$. Each uncertainty signal provides effective feedback guidance and improves performance, while their combination in the full approach yields the best results.

We present the training dynamics of accuracy in Figure~\ref{fig:acc_branch}, comparing dynamic branch prioritization with fixed-probability sampling. We observe that dynamic branch prioritization consistently achieves higher accuracy throughout training. These results highlight its advantage in balancing exploration and stability: increased reliance on the noisy branch facilitates exploration during early training, while progressively emphasizing the raw branch stabilizes optimization and improves convergence in later stages, leading to superior final performance.

\subsection{Evaluation on More Benchmarks} \label{app: add}

We evaluate on additional benchmarks, including ChartMuseum \citep{tang2026chartmuseum}, MMReason \citep{yao2025mmreason}, and VisuLogic \citep{xu2025visulogic}. As shown, DUPL improves the base 7B model by average 4.0\% and outperforms GRPO.

\begin{table}[ht]
\centering
\resizebox{\linewidth}{!}{%
\begin{tabular}{lcccc}
\toprule
\textbf{Model} & \textbf{ChartMuseum} & \textbf{MMReason} & \textbf{VisuLogic} & \textbf{Avg.} \\
\midrule
Qwen2.5-VL-7B & 26.8 & 16.8 & 26.0 & 23.2 \\
\quad + GRPO & 28.8 & 18.9 & 28.5 & 25.4 \\
\quad + DUPL & \bf 31.3 & \bf 20.5 & \bf 29.8 & \bf 27.2 \\
\bottomrule
\end{tabular}
}
\caption{\textbf{Evaluation on additional multimodal reasoning benchmarks.} DUPL improves the 7B base model by an average of 4.0\% and outperforms the GRPO baseline.}
\label{tab:additional_benchmarks}
\end{table}

\subsection{Statistical Significance} \label{app: seed}

We train the models across four random seeds and evaluate on six benchmarks, reporting the mean and standard deviation in Table~\ref{tab:seed}. DUPL consistently outperforms the 7B base model and the GRPO baseline, confirming both statistical significance and robust training stability.

\begin{table}[ht]
\centering
\footnotesize 
\setlength{\tabcolsep}{6pt} 
\begin{tabular}{lccc} % Only 4 columns now, it will fit perfectly
\toprule
\textbf{Benchmark} & \textbf{Base 7B} & \textbf{GRPO} & \textbf{DUPL (Ours)} \\
\midrule
MathVerse  & 45.8 & 48.1 $\pm$ 0.2 & \textbf{52.2} $\pm$ 0.3 \\
MathVista  & 67.2 & 70.5 $\pm$ 0.4 & \textbf{74.3} $\pm$ 0.2 \\
WeMath     & 63.2 & 68.4 $\pm$ 0.1 & \textbf{71.1} $\pm$ 0.2 \\
HalluBench & 65.2 & 68.7 $\pm$ 0.3 & \textbf{71.2} $\pm$ 0.3 \\
ChartQA    & 79.8 & 81.8 $\pm$ 0.2 & \textbf{83.9} $\pm$ 0.1 \\
LogicVista & 45.5 & 46.2 $\pm$ 0.3 & \textbf{48.8} $\pm$ 0.2 \\
\midrule
Average & 61.1 & 64.0 & \textbf{66.9} \\
\bottomrule
\end{tabular}
\caption{\textbf{Evaluation on six benchmarks across four random seeds}. We report mean and standard deviation. DUPL consistently improves over GRPO and the base model, demonstrating both performance gains and statistical significance.}
\label{tab:seed}
\end{table}

\subsection{Evaluation with Qwen3-VL} \label{app: qwen3}

We evaluate our framework using Qwen3-VL-Instruct 4B and 8B models. As shown in Table \ref{tab:qwen3},  DUPL improves accuracy by up to 10.7\% (Avg.\ 5.3\%) for the 4B base model and up to 12.4\% (Avg.\ 6.3\%) for the 8B base model across all evaluated tasks. Furthermore, DUPL consistently outperforms the strong GRPO baseline, with DUPL-8B achieving the best overall average performance.

\begin{table*}[t]
    \centering
    \resizebox{\textwidth}{!}{
    \begin{tabular}{lcccccc}
        \toprule
        \textbf{Model} & \textbf{MathVerse} & \textbf{MathVista} & \textbf{WeMath} & \textbf{HalluBench} & \textbf{ChartQA} & \textbf{Avg.} \\
        \midrule
        Qwen3-VL-4B-Instruct \citep{bai2025qwen3} & 57.6 & 75.5 & 65.9 & 73.4 & 79.5 & 70.4 \\
        \quad + GRPO & 62.0  & 78.2 & 76.3 & 74.7 & 81.7 & 74.6 \\ 
        \rowcolor{gray!15} \quad + DUPL (Ours) & \bf 65.1 & \bf 78.4 & \bf 76.6 & \bf 75.7 & \bf 82.8 & \bf 75.7 \\
        \midrule
        Qwen3-VL-8B-Instruct \citep{bai2025qwen3}& 58.6 & 76.8 & 70.5 & 74.4 & 79.6 & 72.0 \\
        \quad + GRPO & 63.4 & 78.9 & 77.6 & 75.6 & 82.2 & 75.5 \\ 
        \rowcolor{gray!15} \quad + DUPL (Ours) & \bf 68.1 & \bf 80.0 & \bf 82.9 & \bf 76.6 & \bf 84.1 & \bf 78.3 \\
        \bottomrule
    \end{tabular}
    }
    \vspace{-5pt}
        \caption{\textbf{Evaluation performance using Qwen3-VL models.} Compared to the base models, DUPL improves accuracy by up to 10.7\% (Avg.\ 5.3\%) for the 4B model and up to 12.4\% (Avg.\ 6.3\%) for the 8B model across all evaluated tasks. Furthermore, DUPL consistently outperforms the strong GRPO baseline, with DUPL-8B achieving the best overall average performance.}
    \label{tab:qwen3}
    \vspace{-12pt}
\end{table*}

\end{document}